\pdfoutput=1
\documentclass[11pt]{article}
\usepackage[final]{acl}
\usepackage{times}
\usepackage{latexsym}
\usepackage[T1]{fontenc}

\usepackage[utf8]{inputenc}
\usepackage{microtype}
\usepackage{inconsolata}

\usepackage{graphicx}
\usepackage{amsmath}
\usepackage{url}
\usepackage{comment}

\usepackage{algorithm}
\usepackage{algorithmic}

\usepackage[group-separator={,}]{siunitx}
\usepackage{hyperref}

\usepackage[super]{nth}

\title{Identifying Emerging Concepts in Large Corpora}

\author{
    Sibo Ma \hspace{0.4cm} Julian Nyarko \\
    Stanford University \\
    \texttt{siboma@stanford.edu} \hspace{0.4cm} \texttt{jnyarko@stanford.edu}
}

\date{\today}

\begin{document}
\maketitle

\begin{abstract}
We introduce a new method to identify emerging concepts in large text corpora. By analyzing changes in the heatmaps of the underlying embedding space, we are able to detect these concepts with high accuracy shortly after they originate, in turn outperforming common alternatives. We further demonstrate the utility of our approach by analyzing speeches in the U.S. Senate from 1941 to 2015. Our results suggest that the minority party is more active in introducing new concepts into the Senate discourse. We also identify specific concepts that closely correlate with the Senators' racial, ethnic, and gender identities. An implementation of our method is publicly available.
\end{abstract}

\section{Introduction}
The identification of new ideas and concepts within large corpora is of core interest both in computational linguistics, the social sciences, and the humanities. For instance,~\citet{hofstra-science} discover new innovations in a corpus of scientific articles. They find that minorities often introduce novel contributions to the scientific discourse, but those innovations are disproportionately discounted. \citet{Tessa-social-embeddings} examine how new stereotypes towards ethnic and racial minorities evolved during the \nth{19} and \nth{20} centuries. And~\citet{hanley2024specious} identify misinformation at its conception and track its spread and influence on public discourse.

Recent advances in natural language processing offer more robust alternatives. Transformer-based models, such as BERT~\citep{devlin-etal-2019-bert} and T5~\cite{Raffel-2020}, leverage contextual embeddings to encode semantic meaning beyond individual words, making them well-suited for tracking emerging concepts that lack stable lexical forms. However, most existing methodologies for text analysis are not specifically designed for identifying emergent concepts. Instead, they often apply general-purpose techniques that do not account for the distinct temporal patterns associated with conceptual emergence. 

As a consequence, existing approaches often lack sensitivity, allowing for the accurate identification of emerging concepts only after these concepts have become well-represented in the underlying corpora. To the extent that methods exist that are specifically targeted at emerging concept identification, they tend to be supervised~\cite{Tessa-social-embeddings,Kulkarni-linguistic-change}, thus requiring the investigator to know \textit{ex ante} what new concepts to look for. They also tend to operate at the word-level, thus making it impossible to identify concepts that are not easily representable using a distinct unigram or bigram. But consistent with the semantic perspective, we understand concepts as abstract objects—propositions that can be expressed in a multitude of ways~\cite{margolis2007ontology}. While some concepts can be captured by individual words or phrases (e.g., ``climate change''), we assume that many concepts are more complex and cannot be reduced to a single lexical unit (e.g., ``negative sentiment towards the Affordable Care Act as a form of governmental invasion''~\cite{fisher2019virtual}). This in turn requires a methodological approach that goes beyond the word-level and is able to capture the broader semantic structures that define emerging concepts.

In this paper, we introduce a novel methodology to identify emerging concepts in large text corpora that we also make publicly available.\footnote{An implementation can be found at \url{https://github.com/Crabtain959/new_concept_detection}.} 
Intuitively, our methodology leverages heatmaps of the embedding space to identify those regions that are subject to sudden, long-lasting increases in density. Because we benchmark the density increase against commonly observed, non-systematic changes, our method performs particularly well at identifying emerging concepts shortly after their inception, well before they become broadly represented in the underlying corpus. Our method operates at the sentence level (or any other, larger textual unit), thus allowing for identification of complex concepts that cannot easily be captured with a distinct word or phrase.

In several evaluations, we demonstrate the performance of our proposed methodology, and compare it to other approaches that have been employed to identify emerging concepts in text. In a last step, we illustrate the utility of our method for the social sciences and digital humanities by exploring the introduction of new concepts in the U.S. Senate debates from 1941 (\nth{77} Congress) to 2015 (\nth{114} Congress). At a macro level, we find a consistent pattern showing that the minority party introduces new concepts and ideas with greater frequency than the majority. This finding lends support to claims by other scholars about strategically different behavior of the minority party in legislative bodies~\cite{Jenkins_Monroe_Provins_2023, BALLARD_CURRY_2021}, including their way to converse~\cite{Pozen_Talley_Nyarko_2019}.

\section{Related Work}
Perhaps closest in spirit to our motivation is a recent study by~\citet{idea-emerge}. They identify novel ideas using word-level perplexity, such that sentences with unexpected word combinations are deemed to reflect novelty. They provide convincing evidence that their approach identifies novel ideas at a macro level. In their approach, syntactic idiosyncracies are receiving the same weight as unusual semantic occurrences. And since it operates at the word level, there are likely limits to the linguistic complexity with which new ideas can be identified.

In contrast, our approach centers around textual embeddings, a literature with a long tradition. A number of early contributions interested in the identification of new concepts characterized the problem in terms of diachronic meaning shift detection. That is, a new concept would be identified by the emergence of a new word sense for a given word. Notably,~\citet{hamilton-etal-2016-diachronic} demonstrated that word meaning evolution follows predictable trajectories, influenced by frequency and semantic drift, highlighting the importance of tracking concept emergence over time. 
Naturally, the associated methodologies are centered around the use of word embedding models. Several studies have explored different approaches to detecting semantic change, including cosine distance between word embeddings~\citep{del-tredici-etal-2019-short, Kulkarni-linguistic-change, shoemark-etal-2019-room}, Bayesian models for temporal word representations~\citep{frermann-lapata-2016-bayesian}, and topic modeling techniques such as Latent Dirichlet Allocation (LDA)~\citep{lau-etal-2012-word, lda}. Other refinements incorporate contextual embeddings~\citep{martinc-etal-2020-leveraging} or focus on modeling semantic drift in historical corpora~\citep{periti2024studying, boholm2024can}. Surveys such as~\citet{semanticsurvery2024} provide an overview of these methodologies and their relative strengths and limitations.

Despite their effectiveness in tracking known shifts, these methods face two key limitations. First, since they analyze shifts at the word-level, they are not well-suited to detect emerging concepts that cannot easily be associated with a single word or phrase~\citep{stewart-eisenstein-2018-making, hofmann-etal-2020-predicting}. Second, these methods are supervised in the sense that the investigator needs to predefine the list of words for which a new meaning may emerge. In that way, they are better suited to identify \textit{when} a known semantic shift has occurred. However, they are not aimed at identifying previously unknown, emergent concepts.

More recent approaches have relied on the detection of concepts by applying clustering algorithms to text embeddings~\cite{sia2020tired,giulianelli-etal-2020-analysing,angelov2020top2vec,grootendorst2022bertopic}. These unsupervised methods do not require predefined knowledge about the changing concepts, making them better-suited to detect unknown concepts in large corpora. However, clustering methods are insensitive to the temporal nature of the data generating process, thus preventing them from taking into account the dynamic features that accompany an emergent concept. In addition, the most commonly employed clustering methods like HDBSCAN~\cite{ricardo2013hdbscan} and KMeans variants~\cite{sia2020tired, IKOTUN2023178} suffer from parameter selection and difficulty in controlling sensitivity and specificity~\cite{hanley2024specious}. For instance, the widely used HDBSCAN algorithm~\cite{ricardo2013hdbscan} (see, e.g. ~\citealp{grootendorst2022bertopic}) is very sensitive to its core parameters, including the minimum number of points required to form a cluster, the minimum number of samples in a neighborhood for a point to be considered a core point, and the distance threshold for merging clusters. Only most recently have researchers begun to adapt these approaches by examining the temporal or usage changes within clusters~\cite{hanley2024specious}, but the current clustering methods lack the sensitivity to capture more nuanced topics. We will further evaluate this in section~\ref{section:synthetic}.

Our method was inspired by these clustering approaches and aims to improve upon them by using blob detection with Laplacian of Gaussian Filtering~\cite{6408211} on differences of heatmaps. This technique mitigates the complexity of parameter selection and enhances control over robustness and sensitivity. Instead of clustering all data points across time periods together and analyzing data composition within each cluster, we track the development of concepts by detecting and examining regions of high, novel density in the embedding space. By focusing on differences between time periods, this approach removes noise and allows for more accurate and sensitive detection of emerging concepts.

\section{Method}

This section outlines our method to detect emerging concepts. In doing so, we follow~\citet{idea-emerge} and define new linguistic concepts as those that emerge as separate and distinct from existing discourse and remain with some permanence.

\begin{figure*}[t]
    \centering
    \includegraphics[width=\linewidth]{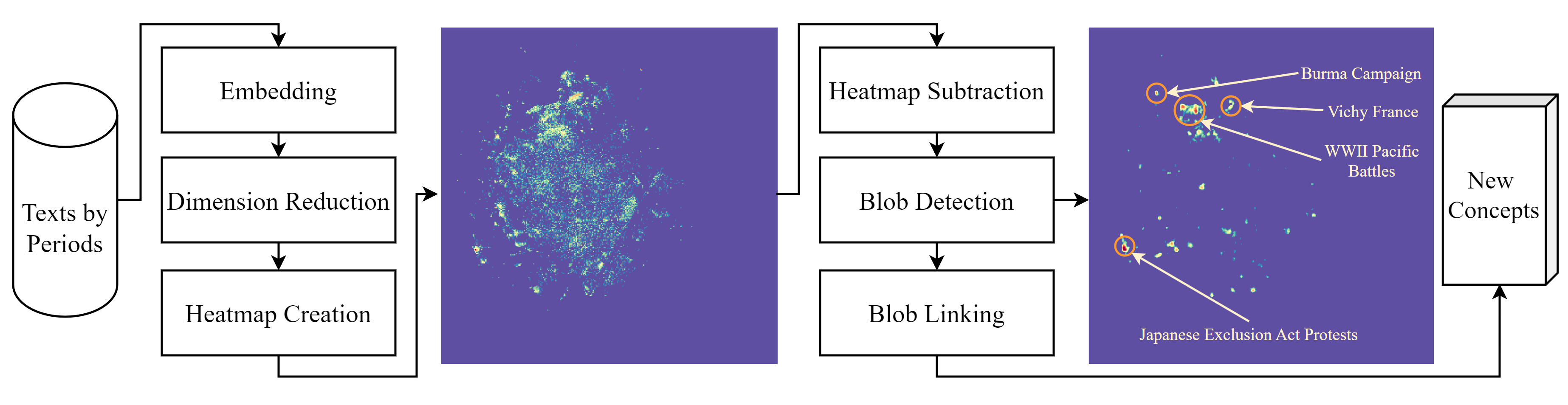}
    \caption{Overview of our approach for detecting new topics. Texts are embedded and processed into lower-dimensional heatmaps. The heatmap on the left visualizes an example distribution of embeddings after dimensionality reduction. Next, heatmap subtraction removes existing patterns, leaving new regions of high density. The heatmap on the right shows the embeddings after subtraction, with blobs representing new concepts. These blobs are then detected and linked to form cohesive new concepts, which are labeled in the final output (e.g., Burma Campaign, WWII Pacific Battles, Japanese Exclusion Act Protests)}
    \label{fig:mesh1}
\end{figure*}

Our process involves several key steps: (1) embedding sentences to capture their semantic features, (2) reducing the dimensionality of these embeddings and generating heatmaps to summarize the embedding space and its distribution, (3) detecting significant changes among the heatmaps with blob detection on the differences among heatmaps, and (4) tracking the progression of these changes and interpreting them as newly emerging concepts. Each step is designed to ensure that the analysis is both comprehensive and efficient, and gives informative results for downstream analysis. Figure~\ref{fig:mesh1} illustrates our pipeline, which is explained in more detail in the following subsections.

\subsection{Sentence Embedding and Dimensionality Reduction}
We use the MPNet model~\cite{song2020mpnetmaskedpermutedpretraining} to embed all the sentences, capturing their semantic features\footnote{When this study started in 2023, MPNet was the most suitable and well-performing model for our task. It ranked among the top models for MTEB Clustering~\cite{muennighoff-etal-2023-mteb}, alongside ST5-XXL. Given our objective of ensuring broad accessibility and usability of our pipeline for social scientists, we selected MPNet over ST5-XXL, as it provides comparable performance while being substantially more computationally efficient.}. 

To utilize the resulting embeddings for the creation of a heatmap, we first reduce their dimensions via Uniform Manifold Approximation and Projection (UMAP) \cite{mcinnes2020umap} with default parameters\footnote{During early experimentation, we found that final results are not very sensitive to the parameters. Especially after subtracting heatmaps, different parameters gave difference heatmaps with similarly clear patterns for blob detection. Since the cost-performance ratio was not high, we decided to illustrate our pipeline with the default parameters. A robustness test with different parameters can be found in~\autoref{sec:robustness-test}, yielding similar results}. We opt for UMAP due to its ability to maintain the global topological structure of the data while reducing its dimensionality. It operates by constructing a high-dimensional graph representation of the data, which is then approximated in a lower-dimensional space. UMAP optimizes a cost function which is based on a cross-entropy between the distances in the high-dimensional space and their representation in the low-dimensional space. This approach allows UMAP to balance attention between local and global structures, ensuring that similar points remain close to each other in the reduced space while also appropriately modeling the broader dataset topology.

To further ensure embeddings are mapped consistently, we fit the UMAP on all the sentence embeddings, ensuring that the topological structure is maintained and thus guarantees that embeddings will be mapped consistently. This consistency is crucial for subsequent analysis and comparison because we will create heatmaps and compare absolute data positions across different periods. 

We choose the target dimension of $n=2$, which minimizes memory and computation requirements while still providing results that are informative enough. We discuss the tradeoff between computational efficiency and the richness of the representation in more detail in the next section. 

\subsection{Heatmap Generation}
For each time period, we take the reduced-dimension embeddings of each year and create n-dimensional histograms as heatmaps. These heatmaps represent the density and distribution of sentence embeddings over time in a compact manner, which serves as an initial stage of generalization.

The creation of the heatmaps follows the following process:  
\begin{enumerate}
    \item \textbf{Define the Range:} Set the range of each dimension from the minimum to the maximum value of all embeddings in that dimension.
    \item \textbf{Create Bins:} Split each dimension into $m$ parts, resulting in $m^n$ bins. We choose $m=400$ to balance granularity and generalization, resulting in $400^2=160000$ bins in our settings.
    \item \textbf{Fill Bins:} For each embedding, determine the corresponding bin based on its coordinates and count the number of embeddings in each bin to determine the density.
\end{enumerate}

If a new concept appears in period $p$, it should not be significantly present before $p$, but should be prominent in and after $p$ for a certain number of periods. Our parameters are designed to accommodate concepts with varying characteristics (e.g., duration of prominence), and their specific choices are detailed in~\autoref{sec:paramter-choice}. To capture this temporal change, we first select a set of $R$ reference heatmaps $M_{p-r}$ $(1 \leq r \leq R)$ from periods preceding $p$. The reference heatmaps are summarized by taking, for each bin, the maximum value across $M_{p-r}$, denoted as $RM_p$. By imposing constraints on $RM_p$, we can guarantee that a new concept has not been discussed frequently in any of the previous periods.

\begin{equation}
    \label{eq:ref-heatmap}
    RM_p(i, j) = \max_{1 \leq r \leq R} M_{p-r}(i, j),
\end{equation}

To mitigate the influence of different density scales across periods, we normalize each heatmap $M_{p+w}$ with respect to the reference heatmap $RM_p$. We then subtract $RM_p$ from the heatmaps $M_{p+w}$ $(0 \leq w < W)$ corresponding to the subsequent $W$ periods. The normalization and subtraction are shown in Equation~\ref{eq:heatmap}. This subtraction highlights variations and shifts in sentence embeddings, effectively illustrating changes in semantic content over time within the window of $W$ periods. To focus on emerging patterns, we set negative values to zero, as they reflect disappearance. The resulting heatmaps, termed difference heatmaps $DM_{p,w}$, emphasize regions with significant changes relative to the summarized reference periods $RM_p$. 

\begin{equation}
  \label{eq:heatmap}
  DM_{p,w} = \max \left( 0, \frac{\sum_{e \in RM_p} e}{\sum_{e \in M_{p+w}} e} M_{p+w} - RM_p \right)
\end{equation}

\subsection{Blob Detection}

Next, we identify emerging concepts using blob detection. We use the Laplacian of Gaussian (LoG) method~\cite{6408211} to detect blobs\footnote{We use the python package \textit{scikit-image} to detect blobs~\cite{van2014scikit}.} on the difference heatmaps $DM_{p,w}, 1\leq w \leq W$, for each reference period $p$. Blobs, which are regions in the difference heatmaps where significant changes in density occur, indicate new concepts emerging after reference period $p$.

Technical Details on LoG when $DM_{p,w}$ is 2-dimensional:
\begin{enumerate}
    \item \textbf{Gaussian Kernel:} 
    
    The Gaussian kernel is used to smooth the difference heatmaps $DM(x,y)$. It is defined as the following:
    \begin{equation}
        G(x,y;\sigma) = \frac{1}{2\pi\sigma^2}e^{-\frac{x^2+y^2}{2\sigma^2}}
    \end{equation}
    
    where, $x$ and $y$ are spatial coordinates, and $\sigma$ represents the standard deviation of the Gaussian kernel, which controls the smoothing strength. Larger values of $\sigma$ produce stronger smoothing, affecting a broader area around each bin. 
    \item \textbf{Scale-Space Representation:}
    
    The difference heatmaps $DM(x,y)$ are convolved with the Gaussian kernel $G(x,y;\sigma)$ to produce a scale-space representation:
    \begin{equation}
        L(x,y;\sigma) = G(x,y;\sigma) * DM(x,y)
    \end{equation}
    The result, $L(x,y;\sigma)$, is referred to as the scale-space representation of the difference heatmap $DM(x,y)$. This operation blends the bin values with the degree of blending defined by the Gaussian kernel. It reduces variations and noise in the difference heatmaps, making the resulting heatmaps more robust to noise.
    \item \textbf{Laplacian Operator:} 
    
    The Laplacian operator is applied to the scale-space representation to identify regions where the intensity changes rapidly. The Laplacian operator is a second-order differential operator in two dimensions, calculated as the sum of the second derivatives of $L$ with respect to $x$ and $y$ (noted as $L_{xx}$ and $L_{yy}$, respectively). This operator measures the rate at which the first derivatives change, providing a way to capture regions of rapid intensity change in the image, which often correspond to edges or, as is relevant here, blob-like structures.
    \begin{equation}
        \nabla^2L = L_{xx}+L_{yy}=\frac{\partial^2L}{\partial x^2}+\frac{\partial^2L}{\partial y^2}
    \end{equation}

    \item \textbf{Blob Detection:} Local maxima above a calculated minimum peak intensity in the Laplacian response are identified as blobs. The minimum peak intensity is determined by multiplying the maximum intensity by a relative threshold, referred to as $\rho^*$. The threshold ensures that only prominent features are detected as blobs, allowing for robust detection of important structures.
\end{enumerate}

\subsection{Blob Linking}
To track the development of detected blobs over time, we link blobs across different years, forming a temporal graph. For each blob $b_{p,w,i}$ in a period $w$ with a reference period $p$, we identify blobs $b_{p,w-q,j}$, with $1\leq q \leq Q$, in the earlier period that are within a certain threshold of distance, connect them, and add $b_{p,w,i}$ to the graphs that end with $b_{p,w-q,j}$. If no close blobs in the earlier period are found, we initialize a graph with $b_{p,w,i}$ being the start. This process creates a list of networks of blobs for each reference period, grouping sentences with similar semantic features and showing the progression and transformation of significant semantic regions over time. The pseudo-code for blob linking is presented in Algorithm~\ref{code-blob-linking}.

\subsection{Parameter Choice}
\label{sec:paramter-choice}
Our parameters are set to identify suddenly emerging, long-living concepts under the assumption that those are the most significant. However, those with interest in identifying other temporal patterns may opt for a different set of parameters. For instance, researchers who are interested in including faddish concepts into the analyses may choose to set the window size to a small value (e.g. $W=1$ or $W=2$). Separately, an investigator might be interested in identifying concepts that are being rediscovered. This can be achieved by using a large window size $W$ and a large blob distance $Q$ (e.g. setting $W=30$ and $Q=20$ to detect concepts that appeared within 30 periods and but across up to 20 periods). More generally, blob linkage and parameter choice allow our approach to be flexibly adjusted to identify various temporal patterns. Examples of faddish concepts and rediscovered concepts are included in~\autoref{sec:fad}.

Beyond these task-dependent parameters, practical applications often involve limited prior knowledge about the dataset or the specific concepts of interest. A common strategy, therefore, is to detect concepts over-inclusively and to then filter out false positives. For example, as discussed in~\autoref{section:coha} and~\autoref{section:application}, we choose a small value $\rho^*=0.05$ to permissively detect the blobs. The blob linking step helps mitigate noise by considering only those blobs that appear in similar regions across multiple periods. Further post-processing can also refine results by filtering linked blobs based on additional criteria, such as only keeping the linkage of blobs with a relatively large threshold of number of sentences when one is interested in concepts that are relatively popular.

\section{Evaluation}

We next turn to evaluating our algorithm, contrasting its performance to the popular alternatives introduced by~\citet{grootendorst2022bertopic} and~\citet{hanley2024specious}. A holistic evaluation of new concept detection is inherently difficult and prohibitively costly, for at least three reasons: First, there is little unanimity or consistency in defining the outer contours of a concept. Two experts might disagree, for instance, on whether the Pacific Battle during World War 2 is a separate concept from the attack on Pearl Harbor, or whether one is a smaller concept inside the other. Second--and relatedly--validating the existence of an emerging concept might require extensive domain expertise. Prior papers have often validated their approaches using historical corpora~\cite{idea-emerge,giulianelli-etal-2020-analysing}, but it can be difficult for an untrained human evaluator to identify complex and nuanced concepts without sharp contours, such as those promoting economic mobility.\footnote{For instance, our later analysis reveals emergent concepts in Senate debates surrounding the development of the workforce through trainings and education within economically disadvantaged communities.} In effect, this means that the generation of human labels can be prohibitively costly, especially in contexts where domain expertise is required. Third, creating a comprehensive list of all concepts in any given domain is infeasible due to the sheer number of conceptual developments. This makes it impossible for researchers to assess Recall with any significant reliability.

Although it is not possible for us to remedy these shortcomings directly, we are taking a multi-pronged approach to alleviate concerns as much as possible. In doing so, we take the following steps:

Our first evaluation uses synthetic data. In the synthetic dataset, we are able to randomize the baseline corpus, introducing new concepts artificially in a controlled way. Randomization ensures that the baseline corpus--in expectation--does not include any coherent concepts, thus allowing us to evaluate both Precision and Recall of our approach. That said, a synthetic corpus has the shortcoming that it does not represent documents that were created through a realistic data generating process, threatening external validity of the results.

We thus complement the evaluation on a synthetic corpus with an assessment using real data, specifically the Corpus of Historical American English (COHA)~\cite{COHA}. As pointed out above, COHA does not allow us to holistically evaluate Recall because no dataset exists that contains the complete set of historical concepts. In addition, we cannot evaluate Precision holistically because there are simply too many concepts to have each verified by domain experts. We thus limit our evaluation on the real data to the identification of a limited set of concepts that do not require extensive domain knowledge for evaluation: Important world events.

\subsection{Synthetic Dataset}
\label{section:synthetic}
We start by evaluating our pipeline on a synthetic dataset. To that end, we collected a number of keywords contained in the WikiPSE dataset~\cite{yaghoobzadeh-etal-2019-probing} and fulfill three conditions: (1) The word had already acquired a distinct meaning in 1910, (2) since 1910, the word acquired an additional, new meaning, and (3) the word is well-represented in COHA ($N\geq500$). For instance, the word \textit{mouse} did historically describe a small rodent, but in 1964, acquired a new meaning as a hand-held, pointing computer device. \autoref{tab:sythetic-eval-keywords} contains the full list of keywords.

\begin{figure}[t]
    \centering
    \includegraphics[width=\columnwidth]{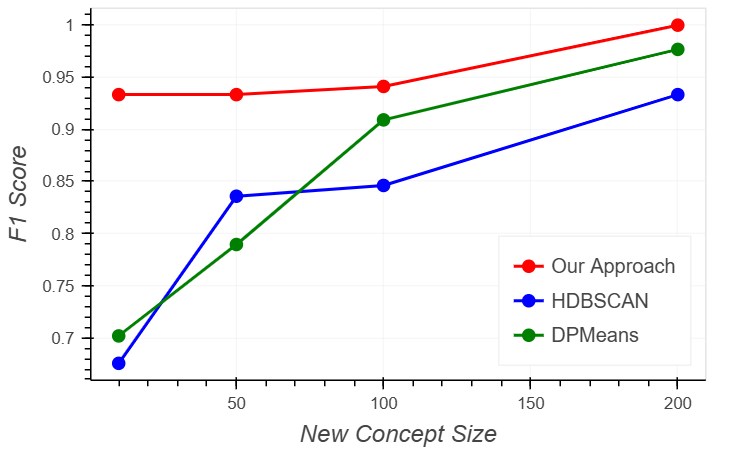}
    \caption{Comparison of $F_1$ scores for three clustering algorithms: Our Approach (red), HDBSCAN (blue), and DPMeans (green) as the size of new topics increases.}
    \label{fig:f1_comparison}
\end{figure}

Next, we create a baseline corpus comprised of \num{207413} texts sourced from newspapers and magazines published between 1900 and 1911, available in COHA. To ensure that this baseline corpus does not contain any temporally correlated, emerging concepts, we then randomize the sentences in our baseline corpus across years. In a next step, we use GPT-4 to generate $n$ sentences containing each selected keyword with their new meanings listed in Table~\ref{tab:sythetic-eval-keywords}\footnote{The prompt for generating the sentences can be found in~\autoref{section:appendix-prompt}.}. The generated sentences are then divided into ten equal sets and introduced into the dataset from 1901 to 1911. Our process ensures that the new concepts in this synthetic dataset exclusively contain, and are limited to, the new meanings of the 8 keywords we manually introduce\footnote{We use 1 NVIDIA A10G GPU for sentence embedding and parallelize the subsequent pipeline steps across 5 AMD EPYC 7R32 CPUs. The entire pipeline completes in 40 minutes, with sentence embedding accounting for the majority of the computational time.}. While the synthetic dataset is constructed around specific keywords, our analysis is conducted at the sentence level, with embeddings capturing broader semantic content beyond individual words. To assess the performance along different sizes of the new topic, we vary $n$ from 10 to 200, treating 1900 as the reference year and applying our pipeline with a window size of $W=10$ and $Q=1$. The parameter $Q$ ensures that only topics present across all 10 periods are detected, aligning with how we construct the synthetic topics.

\autoref{fig:f1_comparison} depicts our performance as $F_1$ scores over the different sizes of new concepts,\footnote{For Recall, a true positive is defined as there being at least one identified new concept with the new meaning of the keyword. For Precision, each new concept with the new meaning of the keyword is a true positive. For example, if the keyword \textit{cool} is split into 2 subconcepts, then they count as 1 true positive for Recall and 2 true positives for Precision.} and compares this performance to the clustering proposals by~\citet{grootendorst2022bertopic} and~\citet{hanley2024specious}. As mentioned above, one disadvantage of these alternatives is their sensitivity to the individual model parameters. In order to avoid biasing results in our favor, we used Bayesian Optimization~\cite{snoek2012practicalbayesianoptimizationmachine} to optimize the model parameters for each approach at each concept size separately.

As can be seen, our pipeline consistently outperforms the two alternatives. The differences are especially pronounced for very small concept sizes. 

This is consistent with our hypothesis that blob detection is effective in capturing temporal changes, as new concepts shortly after inception tend to be smaller and thus more easily detected with our approach

In ~\autoref{fig:parameter-tuning}, we further examine the robustness of our model to the choice of different $\rho^*$, which denotes the threshold for the minimum relative intensity of the peak brightness during blob detection. As can be seen, with $\rho^*\in(0.2,0.4)$, our approach yields good Precision and Recall across different sizes of new topics. 

A qualitative inspection of the results reveals that our approach often further splits keywords into coherent subconcepts. For instance, the keyword \textit{cool} is split into related sub-concepts that describe attire and those that describe behavior, like \textit{cool dance moves}. These first results suggest the pipeline outperforms common alternatives in detecting changes in the semantic landscape, and is able to distinguish between closely related sub-concepts.

\subsection{COHA Dataset}
\label{section:coha}
To validate the adaptability and effectiveness of our pipeline in more complex, real-world scenarios, we extended our analysis to the entire COHA dataset spanning from $1900$ to $2000$ with \num{2870795} sentences.

For each year $p$ from $1900$ to $2000$, we analyze the subsequent $W=10$ of years to detect new concepts that emerged at least twice from $p+1$ to $p+W$\footnote{We set our blob detection parameter to $\rho^*=0.05$ for a more permissive detection as mentioned in~\autoref{sec:paramter-choice}.}.

Given the broad temporal span and the diverse nature of content over a century, our pipeline is expected to capture a wide array of subtle and gradual semantic shifts in this evaluation, which renders it impossible to evaluate Precision or Recall holistically. Due to the absence of a definitive list of emergent concepts over the \nth{20} century, we utilized a chronology of significant political events to assess performance on. The list of historical events used in our analysis and the criteria for their selection are detailed in Appendix~\ref{sec:event-selection}. 

Our pipeline successfully detects all referenced events, though some, such as those related to World War I and World War II, are fragmented into multiple distinct sub-concepts, such as the Pearl Harbor attack and the Burma Campaign. 

\section{Application: Emerging Concepts in the U.S. Congress}
\label{section:application}

\begin{figure}[t]
    \centering
    \includegraphics[width=\columnwidth]{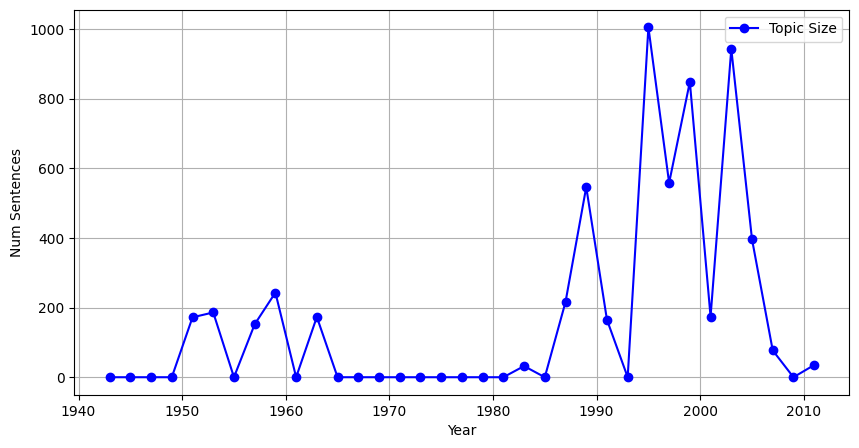}
    \caption{Changes in topic size (number of sentences contained in a topic) over time for Judicial Activism and Marriage Laws. Discussions first emerged in the 1950s and 1960s, with a first major spike in 1989, followed by a series of peaks from 1995 to 2005.}
    \label{fig:topic-evolve}
\end{figure}

\begin{figure}[t]
    \centering
    \includegraphics[width=\columnwidth]{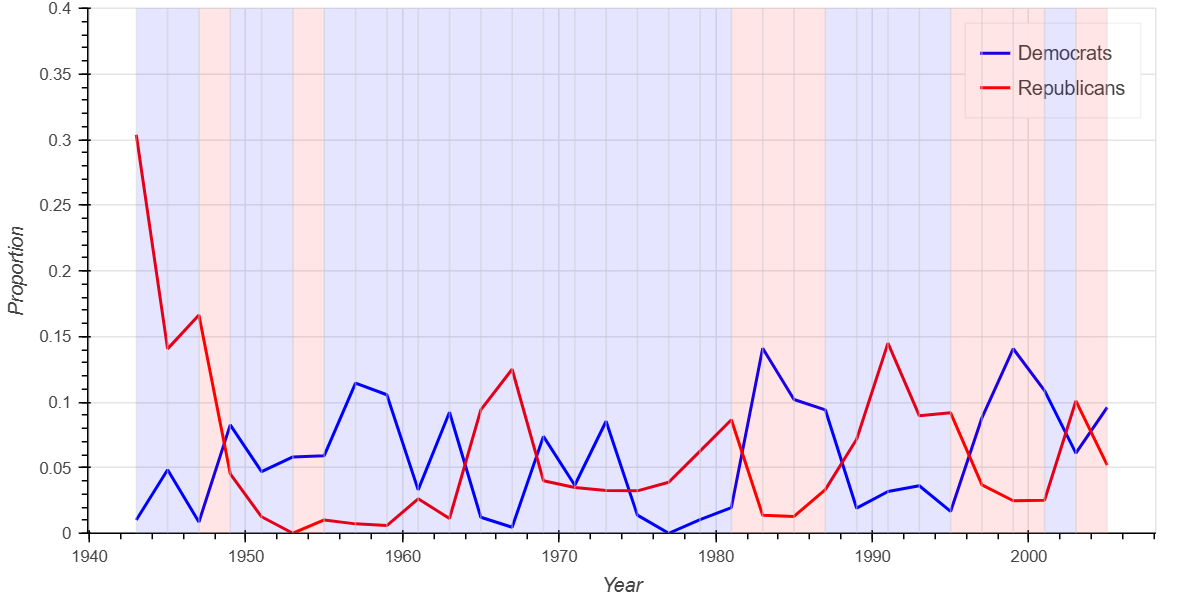}
    \caption{Proportion of new partisan concepts introduced by each party. The red line shows, among all Republican speeches, the proportion of speeches discussing newly introduced, partisan concepts (i.e. concepts for which there is an overrepresentation of Republican speeches). The blue line shows the same for Democratic speeches. The shaded areas indicate periods of party majority: red for Republican majority and blue for Democratic majority.}
    \label{fig:party-talk-innovatively}
\end{figure}

To illustrate the utility of our proposed pipeline in a real-world scenario, we employ it to analyze U.S. Senate speeches derived from the U.S. Congressional Record~\cite{gentzkow_congree} from 1941 (\nth{77} Congress) to 2015 (\nth{114} Congress), which includes a total of \num{2254427} speeches\footnote{We include only speeches from senators and filter out procedural boilerplates, such as expressions of gratitude and requests for unanimous consent, as they do not contribute meaningful content to the concepts of interest}. Our method uncovers emergent concepts and patterns that are not readily detectable using conventional approaches that predominantly focus on syntactic feature extraction. We analyze these concepts both at the macro level by party ideology and at the micro level by Senator identity. As such, our findings contribute to a broader literature on viewpoint diversity at the intersection of NLP and politics~\citep{fridkin2014gender, paul2010summarizing, nemeth2023scoping}.

\autoref{fig:mesh0} provides summary statistics on ideological, gender, and racial/ethnic identity representation, both in terms of personnel and in terms of speech, during our period of observation. These show that, unsurprisingly, the proportion of speeches closely tracks representation in Congress. Interestingly, the results show that women senators initially only rarely spoke in the Senate. Indeed, from 1970 to 1990, the proportion of such speeches was close to 0, although the share of women senators increased steadily over that time period. Even today, women speak disproportionately less in the Senate than men. We observe a similar trend for racial minority senators, which we define as senators with Asian, Black, Hispanic, Native American, or Pacific Islander identity.

\subsection{Illustration of Topic Evolution and Concept Types}
\label{sec:fad}

An illustration of concept evolution using our pipeline is shown in~\autoref{fig:topic-evolve}. In the topic we detected, discussions on Judicial Activism and Marriage Laws first appeared in the 1950s–1960s. The first major spike in 1989 suggests increased legal debates on relationship recognition. The 1995–2005 surge reflects growing attention to marriage definitions and judicial influence.

To detect faddish and rediscovered concepts, we configure parameters as described in~\autoref{sec:paramter-choice}, setting $W=1$ for short-lived fads and $W=30$, $Q=20$ for rediscovered topics. We identified a fad in 1943 on agricultural labor deferments, debating farmworker exemptions from military service amid WWII labor shortages. Meanwhile, a rediscovered topic on employment discrimination emerged in 1961, resurfacing in 1987 and 2013. The sentences had several focal points, including racial equality, responses to legal rulings, and workplace protections for LGBTQ+ individuals. 

\subsection{Minority Party's Innovative Discourse}
At a macro-level, we analyze how the introduction of new, partisan concepts in the Senate correlates with party ideology over time. 

To ensure a focus on substantive concepts, we exclude speeches with an average sentence length of less than 300 characters.\footnote{We set our parameters to $\rho^*=0.05$, consistent with~\autoref{section:coha}. Our parameters $W=10$ and $Q=3$ are more permissive than in our synthetic analysis ($W=10$ and $Q=1$) under the assumption that the underlying data is not as clean as the synthetic data, and so new concepts might disappear intermittently for short periods.} We then compute, for each year,

\begin{equation}
    Q_{p,y} = \frac{\sum_{t \, : \, \frac{S_{p,t}}{S_t} > \frac{N_p}{N}} S_{p,t}}{S_p}
\end{equation}

where $S_{p,t}$ is the number of speeches by party $p$ on concept $t$, $S_{p}$ is the total number of speeches by party $p$, $N_{p}$ is the number of senators from party $p$, $S_{t}$ is the total number of speeches on concept $t$, and $N$ is the total number of senators.

Intuitively, our measure $Q_{p,y}$ captures the proportion of new partisan concepts introduced by each party, ignoring new concepts that do not show a partisan leaning. 

\autoref{fig:party-talk-innovatively} illustrates our findings. Although the initial years do not show a conclusive pattern\footnote{This may, at least in part, be a consequence of the fact that novelty is assessed against prior speeches, and the stock of prior speeches is thin in early years.}, with the beginning of the Civil Rights Era in the 50s and 60s, we observe a trend showing that each parties' Senators become more active in the introduction of partisan concepts when they are in the minority\footnote{Although Senate control shifted multiple times during the 107th Congress (2001–2003), Democrats held the majority for the longest continuous period (June 6, 2001–November 12, 2002). The initial Republican majority (January 20–June 6, 2001) resulted from Vice President Cheney’s tie-breaking vote, while the post-election Republican majority (November 12, 2002–January 3, 2003) was not formally reorganized during
the Senate recess. Given this, we classify the 107th Congress as Democratic majority.}. This is in stark contrast to our descriptive findings in~\autoref{fig:mesh0}, which have shown that the general volume of speeches tracks party representation. In the next subsection, we further break these concepts down by party affiliation, among others.

~\citet{Pozen_Talley_Nyarko_2019} found that Constitutional discourse in Congress is often shaped by the minority. In particular, they suggest that the minority party strategically employs the Constitution to strengthen its arguments against the majority. Our findings, although necessarily limited given their context, lend at least suggestive evidence to the hypothesis that such patterns might characterize the discourse in the Senate in a more fundamental way. In particular, despite speaking less, the minority party appears to use its allotted time strategically to shift the discourse towards new ideas and discourse.

\subsection{New Concepts and Identity}
We complement the preceding macro-level analysis with a micro-level analysis of new concept introduction and ideological, gender, and racial/ethnic identity in the Senate. Specifically, for each Congress, we treat all preceding Congresses as reference periods and analyze the emergence of new concepts in the subsequent 5 Congresses, covering a span of 10 years.\footnote{Due to the large number of concepts, we relied on GPT-4o to generate summaries, which we then checked selectively to confirm accuracy. The prompt we used is presented in \ref{section:appendix-prompt}}

We find that the new concepts with a disproportionate representation of women senators center around concepts such as climate change and environmental policy, health care accessibility, and energy markets. The top 20 detected concepts (based on how strongly they overrepresent women) are included in Section~\ref{section:appendix-women}.

At the same time, racial minority Senators introduce new impulses around the preservation of fundamental benefits like access to healthcare and education for indigenous and marginalized communities, civil rights protections, community safety, and immigration reform. The top 20 detected concepts (based how strongly they overrepresent minorities) are included in Section~\ref{section:appendix-minority}.

Republican senators introduced new concepts around the military and national security, the rising federal debt, and cold war relations with the Soviet Union and Spain, among others. Democratic senators instead set new impulses regarding environmental policy, small business protection, and human rights. The top 20 detected concepts (based on how strongly they overrepresent Republicans and Democrats) are included in Section~\ref{section:appendix-republican} and Section~\ref{section:appendix-democrat}.

\section{Conclusion}
We have introduced a new, unsupervised methodology to identify emerging concepts in large text corpora. Our approach is able to identify new concepts shortly after their inception, before they become deeply entrenched in the discourse. In doing so, we hope our efforts contribute to recent developments that leverage computational linguistics to support new discoveries, especially within the social sciences and digital humanities~\cite{grimmer2022text}. 

\section{Limitations}
Although our method demonstrates high performance in detecting emerging concepts, there are necessarily limitations to our approach. 

For one, although our method relies on an intuitive parameter $\rho^*$, ~\autoref{fig:parameter-tuning} shows that performance can be sensitive to this parameter. To mitigate concerns arising from this sensitivity, we adopt a permissive selection of $\rho^*$, setting it to a low threshold to maximize the capture of potential emerging concepts. While this reduces the risk of missing meaningful patterns, it may also introduce false positives, requiring additional filtering or refinement.
In addition, given the absence of a comprehensive list of concepts in any real-world corpus, our evaluations are limited to assessing either synthetic data (with potentially limited external validity) or a limited notions of recall in real data (our world events). Finally, the concepts our method identifies may require manual review to detect and filter false positives, or to merge conceptual distinctions that are too nuanced for the relevant inquiry. Although tools such as LLMs can be employed to facilitate this task, it may still be associated with significant costs.

\bibliography{references}

\begin{thebibliography}{43}
\providecommand{\natexlab}[1]{#1}

\bibitem[{Angelov(2020)}]{angelov2020top2vec}
Dimo Angelov. 2020.
\newblock \href {https://arxiv.org/abs/2008.09470} {Top2vec: Distributed representations of topics}.
\newblock \emph{Preprint}, arXiv:2008.09470.

\bibitem[{BALLARD and CURRY(2021)}]{BALLARD_CURRY_2021}
ANDREW~O. BALLARD and JAMES~M. CURRY. 2021.
\newblock \href {https://doi.org/10.1017/S0003055421000381} {Minority party capacity in congress}.
\newblock \emph{American Political Science Review}, 115(4):1388–1405.

\bibitem[{Blei et~al.(2001)Blei, Ng, and Jordan}]{lda}
David Blei, Andrew Ng, and Michael Jordan. 2001.
\newblock Latent dirichlet allocation.
\newblock volume~3, pages 601--608.

\bibitem[{Boholm et~al.(2024)Boholm, R{\"o}nnerstrand, Breitholtz, Cooper, Lindgren, Rettenegger, and Sayeed}]{boholm2024can}
Max Boholm, Bj{\"o}rn R{\"o}nnerstrand, Ellen Breitholtz, Robin Cooper, Elina Lindgren, Gregor Rettenegger, and Asad Sayeed. 2024.
\newblock Can political dogwhistles be predicted by distributional methods for analysis of lexical semantic change?
\newblock In \emph{Proceedings of the 5th Workshop on Computational Approaches to Historical Language Change}, pages 144--157.

\bibitem[{Campello et~al.(2013)Campello, Moulavi, and Sander}]{ricardo2013hdbscan}
Ricardo J. G.~B. Campello, Davoud Moulavi, and Joerg Sander. 2013.
\newblock Density-based clustering based on hierarchical density estimates.
\newblock In \emph{Advances in Knowledge Discovery and Data Mining}, pages 160--172, Berlin, Heidelberg. Springer Berlin Heidelberg.

\bibitem[{Charlesworth et~al.(2022)Charlesworth, Caliskan, and Banaji}]{Tessa-social-embeddings}
Tessa E.~S. Charlesworth, Aylin Caliskan, and Mahzarin~R. Banaji. 2022.
\newblock \href {https://doi.org/10.1073/pnas.2121798119} {Historical representations of social groups across 200 years of word embeddings from google books}.
\newblock \emph{Proceedings of the National Academy of Sciences}, 119(28):e2121798119.

\bibitem[{Davies(2022)}]{COHA}
Mark Davies. 2022.
\newblock \href {https://doi.org/11272.1/AB2/EQQKSQ} {{Corpus of Historical American English (COHA)}}.

\bibitem[{Del~Tredici et~al.(2019)Del~Tredici, Fern{\'a}ndez, and Boleda}]{del-tredici-etal-2019-short}
Marco Del~Tredici, Raquel Fern{\'a}ndez, and Gemma Boleda. 2019.
\newblock \href {https://doi.org/10.18653/v1/N19-1210} {Short-term meaning shift: A distributional exploration}.
\newblock In \emph{Proceedings of the 2019 Conference of the North {A}merican Chapter of the Association for Computational Linguistics: Human Language Technologies, Volume 1 (Long and Short Papers)}, pages 2069--2075, Minneapolis, Minnesota. Association for Computational Linguistics.

\bibitem[{Devlin et~al.(2019)Devlin, Chang, Lee, and Toutanova}]{devlin-etal-2019-bert}
Jacob Devlin, Ming-Wei Chang, Kenton Lee, and Kristina Toutanova. 2019.
\newblock \href {https://doi.org/10.18653/v1/N19-1423} {{BERT}: Pre-training of deep bidirectional transformers for language understanding}.
\newblock In \emph{Proceedings of the 2019 Conference of the North {A}merican Chapter of the Association for Computational Linguistics: Human Language Technologies, Volume 1 (Long and Short Papers)}, pages 4171--4186, Minneapolis, Minnesota. Association for Computational Linguistics.

\bibitem[{Fisher and Larsen(2019)}]{fisher2019virtual}
Jeffrey~L Fisher and Allison~Orr Larsen. 2019.
\newblock Virtual briefing at the supreme court.
\newblock \emph{Cornell L. Rev.}, 105:85.

\bibitem[{Frermann and Lapata(2016)}]{frermann-lapata-2016-bayesian}
Lea Frermann and Mirella Lapata. 2016.
\newblock \href {https://doi.org/10.1162/tacl_a_00081} {A {B}ayesian model of diachronic meaning change}.
\newblock \emph{Transactions of the Association for Computational Linguistics}, 4:31--45.

\bibitem[{Fridkin and Kenney(2014)}]{fridkin2014gender}
Kim~L Fridkin and Patrick~J Kenney. 2014.
\newblock How the gender of us senators influences people’s understanding and engagement in politics.
\newblock \emph{The Journal of Politics}, 76(4):1017--1031.

\bibitem[{Giulianelli et~al.(2020)Giulianelli, Del~Tredici, and Fern{\'a}ndez}]{giulianelli-etal-2020-analysing}
Mario Giulianelli, Marco Del~Tredici, and Raquel Fern{\'a}ndez. 2020.
\newblock \href {https://doi.org/10.18653/v1/2020.acl-main.365} {Analysing lexical semantic change with contextualised word representations}.
\newblock In \emph{Proceedings of the 58th Annual Meeting of the Association for Computational Linguistics}, pages 3960--3973, Online. Association for Computational Linguistics.

\bibitem[{Grimmer et~al.(2022)Grimmer, Roberts, and Stewart}]{grimmer2022text}
J.~Grimmer, M.E. Roberts, and B.M. Stewart. 2022.
\newblock \href {https://books.google.com/books?id=dL40EAAAQBAJ} {\emph{Text as Data: A New Framework for Machine Learning and the Social Sciences}}.
\newblock Princeton University Press.

\bibitem[{Grootendorst(2022)}]{grootendorst2022bertopic}
Maarten Grootendorst. 2022.
\newblock \href {https://arxiv.org/abs/2203.05794} {Bertopic: Neural topic modeling with a class-based tf-idf procedure}.
\newblock \emph{Preprint}, arXiv:2203.05794.

\bibitem[{Hamilton et~al.(2016)Hamilton, Leskovec, and Jurafsky}]{hamilton-etal-2016-diachronic}
William~L. Hamilton, Jure Leskovec, and Dan Jurafsky. 2016.
\newblock \href {https://doi.org/10.18653/v1/P16-1141} {Diachronic word embeddings reveal statistical laws of semantic change}.
\newblock In \emph{Proceedings of the 54th Annual Meeting of the Association for Computational Linguistics (Volume 1: Long Papers)}, pages 1489--1501, Berlin, Germany. Association for Computational Linguistics.

\bibitem[{Hanley et~al.(2024)Hanley, Kumar, and Durumeric}]{hanley2024specious}
Hans W.~A. Hanley, Deepak Kumar, and Zakir Durumeric. 2024.
\newblock \href {https://arxiv.org/abs/2308.02068} {Specious sites: Tracking the spread and sway of spurious news stories at scale}.
\newblock \emph{Preprint}, arXiv:2308.02068.

\bibitem[{Hofmann et~al.(2020)Hofmann, Pierrehumbert, and Sch{\"u}tze}]{hofmann-etal-2020-predicting}
Valentin Hofmann, Janet Pierrehumbert, and Hinrich Sch{\"u}tze. 2020.
\newblock \href {https://doi.org/10.18653/v1/2020.acl-main.649} {Predicting the growth of morphological families from social and linguistic factors}.
\newblock In \emph{Proceedings of the 58th Annual Meeting of the Association for Computational Linguistics}, pages 7273--7283, Online. Association for Computational Linguistics.

\bibitem[{Hofstra et~al.(2020)Hofstra, Kulkarni, Galvez, He, Jurafsky, and McFarland}]{hofstra-science}
Bas Hofstra, Vivek~V. Kulkarni, Sebastian Munoz-Najar Galvez, Bryan He, Dan Jurafsky, and Daniel~A. McFarland. 2020.
\newblock \href {https://doi.org/10.1073/pnas.1915378117} {The diversity–innovation paradox in science}.
\newblock \emph{Proceedings of the National Academy of Sciences}, 117(17):9284--9291.

\bibitem[{Ikotun et~al.(2023)Ikotun, Ezugwu, Abualigah, Abuhaija, and Heming}]{IKOTUN2023178}
Abiodun~M. Ikotun, Absalom~E. Ezugwu, Laith Abualigah, Belal Abuhaija, and Jia Heming. 2023.
\newblock \href {https://doi.org/10.1016/j.ins.2022.11.139} {K-means clustering algorithms: A comprehensive review, variants analysis, and advances in the era of big data}.
\newblock \emph{Information Sciences}, 622:178--210.

\bibitem[{Jenkins et~al.(2023)Jenkins, Monroe, and Provins}]{Jenkins_Monroe_Provins_2023}
Jeffery~A. Jenkins, Nathan~W. Monroe, and Tessa Provins. 2023.
\newblock \href {https://doi.org/10.1017/S0143814X2300020X} {Toward a theory of minority-party influence in the u.s. congress: whip counts, amendment votes, and minority leverage in the house}.
\newblock \emph{Journal of Public Policy}, 43(4):722–740.

\bibitem[{Kong et~al.(2013)Kong, Akakin, and Sarma}]{6408211}
Hui Kong, Hatice~Cinar Akakin, and Sanjay~E. Sarma. 2013.
\newblock \href {https://doi.org/10.1109/TSMCB.2012.2228639} {A generalized laplacian of gaussian filter for blob detection and its applications}.
\newblock \emph{IEEE Transactions on Cybernetics}, 43(6):1719--1733.

\bibitem[{Kulkarni et~al.(2015)Kulkarni, Al-Rfou, Perozzi, and Skiena}]{Kulkarni-linguistic-change}
Vivek Kulkarni, Rami Al-Rfou, Bryan Perozzi, and Steven Skiena. 2015.
\newblock \href {https://doi.org/10.1145/2736277.2741627} {Statistically significant detection of linguistic change}.
\newblock In \emph{Proceedings of the 24th International Conference on World Wide Web}, WWW '15, page 625–635, Republic and Canton of Geneva, CHE. International World Wide Web Conferences Steering Committee.

\bibitem[{Lau et~al.(2012)Lau, Cook, McCarthy, Newman, and Baldwin}]{lau-etal-2012-word}
Jey~Han Lau, Paul Cook, Diana McCarthy, David Newman, and Timothy Baldwin. 2012.
\newblock \href {https://aclanthology.org/E12-1060} {Word sense induction for novel sense detection}.
\newblock In \emph{Proceedings of the 13th Conference of the {E}uropean Chapter of the Association for Computational Linguistics}, pages 591--601, Avignon, France. Association for Computational Linguistics.

\bibitem[{Margolis and Laurence(2007)}]{margolis2007ontology}
Eric Margolis and Stephen Laurence. 2007.
\newblock The ontology of concepts-abstract objects or mental representations?
\newblock \emph{No{\^u}s}, 41(4):561--593.

\bibitem[{Martinc et~al.(2020)Martinc, Kralj~Novak, and Pollak}]{martinc-etal-2020-leveraging}
Matej Martinc, Petra Kralj~Novak, and Senja Pollak. 2020.
\newblock \href {https://aclanthology.org/2020.lrec-1.592} {Leveraging contextual embeddings for detecting diachronic semantic shift}.
\newblock In \emph{Proceedings of the Twelfth Language Resources and Evaluation Conference}, pages 4811--4819, Marseille, France. European Language Resources Association.

\bibitem[{Matthew~Gentzkow(2018)}]{gentzkow_congree}
Matt~Taddy Matthew~Gentzkow, Jesse M.~Shapiro. 2018.
\newblock \href {https://data.stanford.edu/congress_text} {Congressional record for the 43rd-114th congresses: Parsed speeches and phrase counts}.

\bibitem[{McInnes et~al.(2020)McInnes, Healy, and Melville}]{mcinnes2020umap}
Leland McInnes, John Healy, and James Melville. 2020.
\newblock \href {https://arxiv.org/abs/1802.03426} {Umap: Uniform manifold approximation and projection for dimension reduction}.
\newblock \emph{Preprint}, arXiv:1802.03426.

\bibitem[{Muennighoff et~al.(2023)Muennighoff, Tazi, Magne, and Reimers}]{muennighoff-etal-2023-mteb}
Niklas Muennighoff, Nouamane Tazi, Loic Magne, and Nils Reimers. 2023.
\newblock \href {https://doi.org/10.18653/v1/2023.eacl-main.148} {{MTEB}: Massive text embedding benchmark}.
\newblock In \emph{Proceedings of the 17th Conference of the European Chapter of the Association for Computational Linguistics}, pages 2014--2037, Dubrovnik, Croatia. Association for Computational Linguistics.

\bibitem[{N{\'e}meth(2023)}]{nemeth2023scoping}
Ren{\'a}ta N{\'e}meth. 2023.
\newblock A scoping review on the use of natural language processing in research on political polarization: trends and research prospects.
\newblock \emph{Journal of computational social science}, 6(1):289--313.

\bibitem[{Paul et~al.(2010)Paul, Zhai, and Girju}]{paul2010summarizing}
Michael Paul, ChengXiang Zhai, and Roxana Girju. 2010.
\newblock Summarizing contrastive viewpoints in opinionated text.
\newblock In \emph{Proceedings of the 2010 conference on empirical methods in natural language processing}, pages 66--76.

\bibitem[{Periti and Montanelli(2024)}]{semanticsurvery2024}
Francesco Periti and Stefano Montanelli. 2024.
\newblock \href {https://doi.org/10.1145/3672393} {Lexical semantic change through large language models: a survey}.
\newblock \emph{ACM Comput. Surv.}, 56(11).

\bibitem[{Periti et~al.(2024)Periti, Picascia, Montanelli, Ferrara, and Tahmasebi}]{periti2024studying}
Francesco Periti, Sergio Picascia, Stefano Montanelli, Alfio Ferrara, and Nina Tahmasebi. 2024.
\newblock Studying word meaning evolution through incremental semantic shift detection.
\newblock \emph{Language Resources and Evaluation}, pages 1--37.

\bibitem[{Pozen et~al.(2019)Pozen, Talley, and Nyarko}]{Pozen_Talley_Nyarko_2019}
David Pozen, Eric~L. Talley, and Julian Nyarko. 2019.
\newblock \href {https://papers.ssrn.com/abstract=3351339} {A computational analysis of constitutional polarization}.
\newblock (3351339).

\bibitem[{Raffel et~al.(2020)Raffel, Shazeer, Roberts, Lee, Narang, Matena, Zhou, Li, and Liu}]{Raffel-2020}
Colin Raffel, Noam Shazeer, Adam Roberts, Katherine Lee, Sharan Narang, Michael Matena, Yanqi Zhou, Wei Li, and Peter~J. Liu. 2020.
\newblock Exploring the limits of transfer learning with a unified text-to-text transformer.
\newblock \emph{J. Mach. Learn. Res.}, 21(1).

\bibitem[{Shoemark et~al.(2019)Shoemark, Liza, Nguyen, Hale, and McGillivray}]{shoemark-etal-2019-room}
Philippa Shoemark, Farhana~Ferdousi Liza, Dong Nguyen, Scott Hale, and Barbara McGillivray. 2019.
\newblock \href {https://doi.org/10.18653/v1/D19-1007} {Room to {G}lo: A systematic comparison of semantic change detection approaches with word embeddings}.
\newblock In \emph{Proceedings of the 2019 Conference on Empirical Methods in Natural Language Processing and the 9th International Joint Conference on Natural Language Processing (EMNLP-IJCNLP)}, pages 66--76, Hong Kong, China. Association for Computational Linguistics.

\bibitem[{Sia et~al.(2020)Sia, Dalmia, and Mielke}]{sia2020tired}
Suzanna Sia, Ayush Dalmia, and Sabrina~J. Mielke. 2020.
\newblock \href {https://doi.org/10.18653/v1/2020.emnlp-main.135} {Tired of topic models? clusters of pretrained word embeddings make for fast and good topics too!}
\newblock In \emph{Proceedings of the 2020 Conference on Empirical Methods in Natural Language Processing (EMNLP)}, pages 1728--1736, Online. Association for Computational Linguistics.

\bibitem[{Snoek et~al.(2012)Snoek, Larochelle, and Adams}]{snoek2012practicalbayesianoptimizationmachine}
Jasper Snoek, Hugo Larochelle, and Ryan~P Adams. 2012.
\newblock Practical bayesian optimization of machine learning algorithms.
\newblock \emph{Advances in neural information processing systems}, 25.

\bibitem[{Song et~al.(2020)Song, Tan, Qin, Lu, and Liu}]{song2020mpnetmaskedpermutedpretraining}
Kaitao Song, Xu~Tan, Tao Qin, Jianfeng Lu, and Tie-Yan Liu. 2020.
\newblock Mpnet: masked and permuted pre-training for language understanding.
\newblock In \emph{Proceedings of the 34th International Conference on Neural Information Processing Systems}, NIPS '20, Red Hook, NY, USA. Curran Associates Inc.

\bibitem[{Stewart and Eisenstein(2018)}]{stewart-eisenstein-2018-making}
Ian Stewart and Jacob Eisenstein. 2018.
\newblock \href {https://doi.org/10.18653/v1/D18-1467} {Making {\textquotedblleft}fetch{\textquotedblright} happen: The influence of social and linguistic context on nonstandard word growth and decline}.
\newblock In \emph{Proceedings of the 2018 Conference on Empirical Methods in Natural Language Processing}, pages 4360--4370, Brussels, Belgium. Association for Computational Linguistics.

\bibitem[{Van~der Walt et~al.(2014)Van~der Walt, Sch{\"o}nberger, Nunez-Iglesias, Boulogne, Warner, Yager, Gouillart, and Yu}]{van2014scikit}
Stefan Van~der Walt, Johannes~L Sch{\"o}nberger, Juan Nunez-Iglesias, Fran{\c{c}}ois Boulogne, Joshua~D Warner, Neil Yager, Emmanuelle Gouillart, and Tony Yu. 2014.
\newblock scikit-image: image processing in python.
\newblock \emph{PeerJ}, 2:e453.

\bibitem[{Vicinanza et~al.(2022)Vicinanza, Goldberg, and Srivastava}]{idea-emerge}
Paul Vicinanza, Amir Goldberg, and Sameer~B Srivastava. 2022.
\newblock \href {https://doi.org/10.1093/pnasnexus/pgac275} {{A deep-learning model of prescient ideas demonstrates that they emerge from the periphery}}.
\newblock \emph{PNAS Nexus}, 2(1):pgac275.

\bibitem[{Yaghoobzadeh et~al.(2019)Yaghoobzadeh, Kann, Hazen, Agirre, and Sch{\"u}tze}]{yaghoobzadeh-etal-2019-probing}
Yadollah Yaghoobzadeh, Katharina Kann, T.~J. Hazen, Eneko Agirre, and Hinrich Sch{\"u}tze. 2019.
\newblock \href {https://doi.org/10.18653/v1/P19-1574} {Probing for semantic classes: Diagnosing the meaning content of word embeddings}.
\newblock In \emph{Proceedings of the 57th Annual Meeting of the Association for Computational Linguistics}, pages 5740--5753, Florence, Italy. Association for Computational Linguistics.

\end{thebibliography}

\appendix
\section{Appendix}
\subsection{Effect of Varying Blob Detection Parameter}
\begin{figure*}
    \centering
    \includegraphics[width=\linewidth]{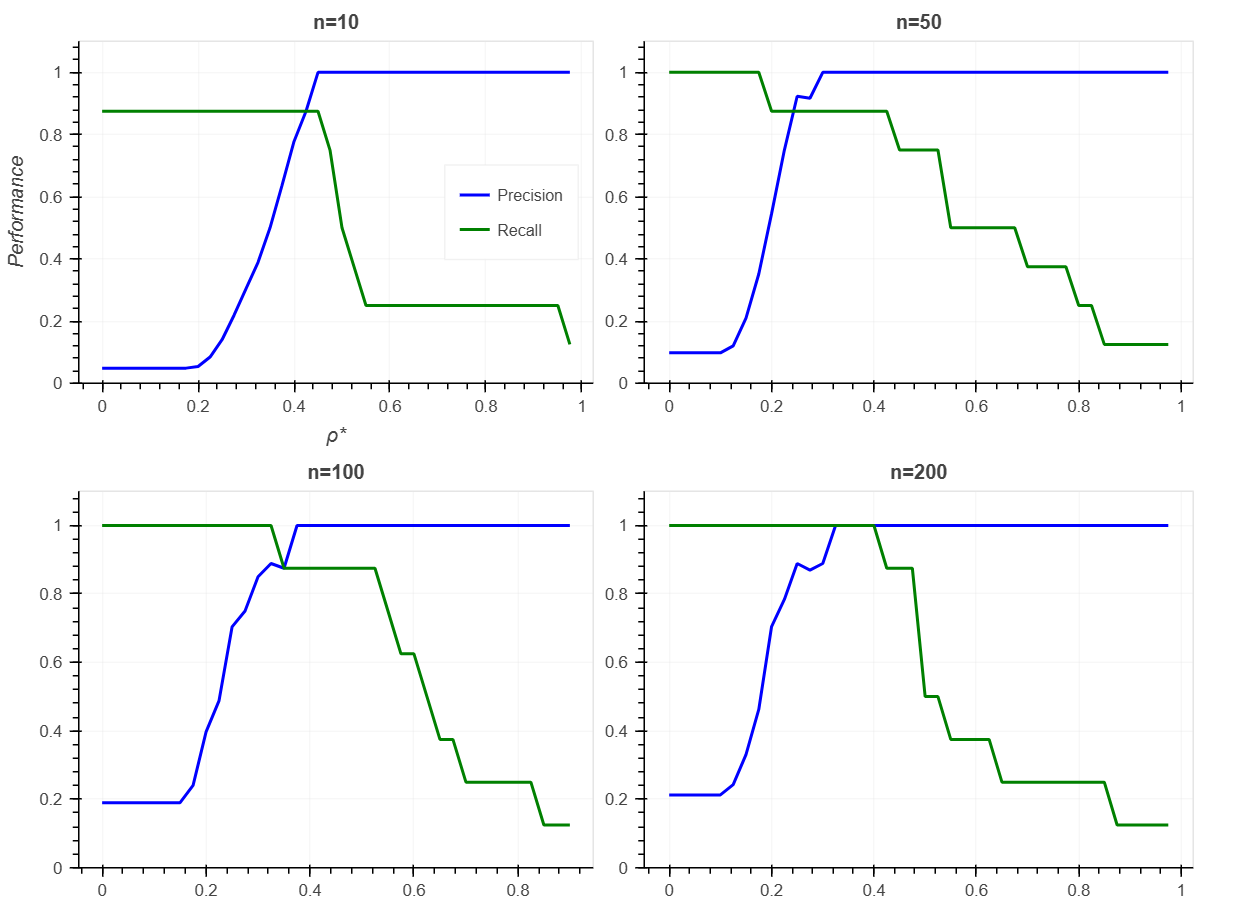}
    \caption{The effect of varying $\rho^*$, the threshold controlling the minimum peak intensity for a blob to be identified, on the pipeline's performance for different sizes of new concepts ($n$). The blue and green lines represent Precision and Recall, respectively.}
    \label{fig:parameter-tuning}
\end{figure*}

\autoref{fig:parameter-tuning} illustrates how different values of $\rho^*$, the threshold controlling the minimum peak intensity for a blob to be identified, affect the pipeline’s performance. The plot examines Precision and Recall across different sizes of new concepts ($n$), with the blue and green lines representing Precision and Recall, respectively. This analysis highlights the trade-off in parameter selection, where lower $\rho^*$ values capture more emerging concepts but may introduce noise, while higher values risk missing smaller but meaningful patterns.

\subsection{Synthetic Evaluation Keywords}
\begin{table*}
\centering
\begin{tabular}{lll}
\hline
\textbf{Keyword} & \textbf{Old Meaning} & \textbf{New Meaning} \\ \hline
Mouse & Small rodent & Computer device \\ 
Gay & Happy or joyous & Homosexual \\ 
Cool & Moderately cold & Stylish or impressive \\ 
Cloud & Mass of condensed water vapor & Online data storage or computing services \\ 
Surf & Riding on the waves on a surfboard & Browse the internet \\ 
Bug & Insect & Software error \\ 
Virus & Infectious biological agent & Malicious software (malware) \\ 
Hack & Cutting with rough blows & Unauthorized access to systems or networks \\ \hline
\end{tabular}
\caption{Old and new meanings of selected keywords}
\label{tab:sythetic-eval-keywords}
\end{table*}

\autoref{tab:sythetic-eval-keywords} presents a selection of keywords used in our synthetic evaluation. 

\subsection{Speech Representation Statistics}
\label{sec:representation}

\begin{figure*}
    \centering
    \includegraphics[width=\linewidth]{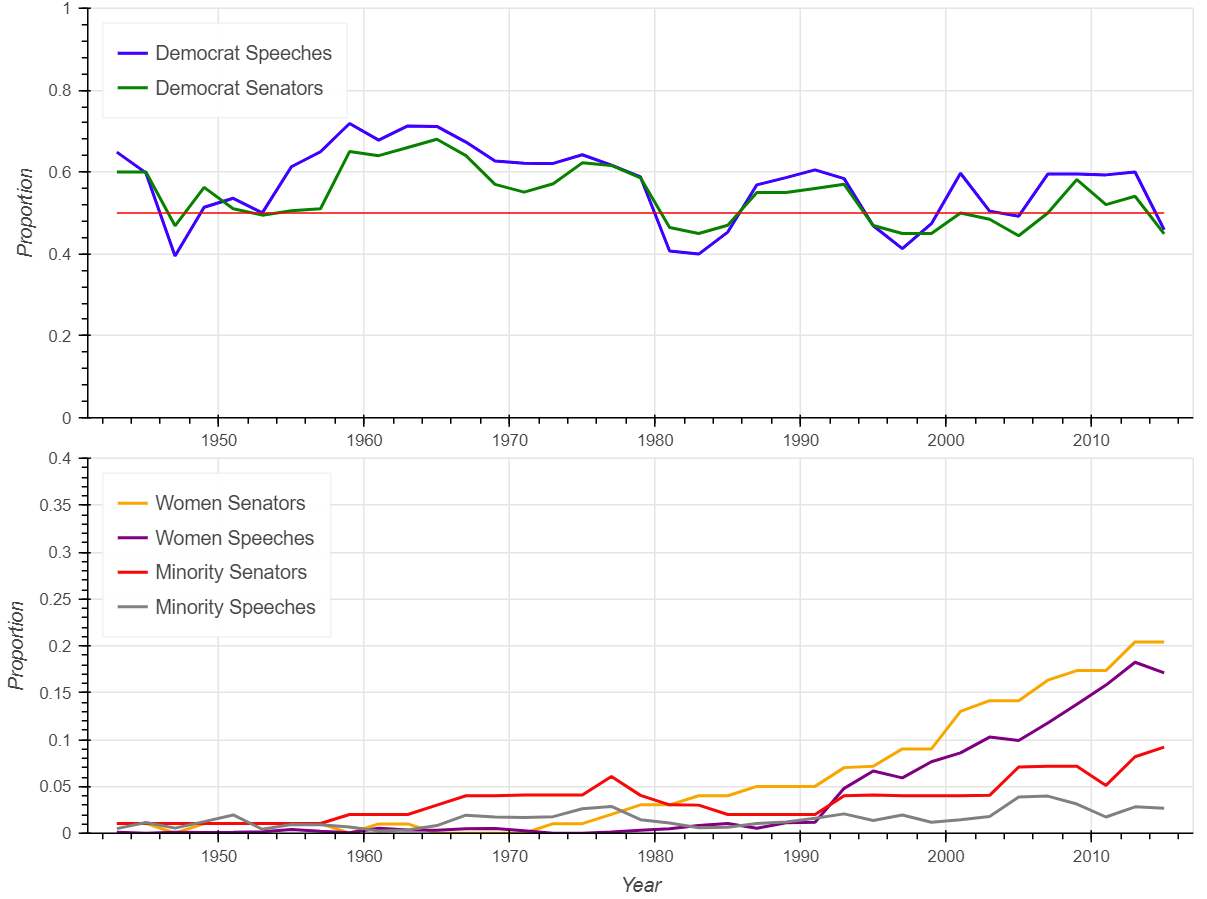}
    \caption{Fraction of speeches in the Congressional Record and fraction of senators by party affiliation, gender, and minority status.}
    \label{fig:mesh0}
\end{figure*}

\autoref{fig:mesh0} provides summary statistics on ideological, gender, and racial/ethnic identity representation, both in personnel and speech, during our period of observation. 

\subsection{Robustness Test}
\label{sec:robustness-test}
\begin{table*}
\centering
\begin{tabular}{lll}
\hline
\textbf{Parameter} & \textbf{Value}       & \textbf{Performance}  \\
\hline

Embedding Model    & stella\_en\_400M\_v5 & 0.94                  \\
UMAP n\_components & 3                    & 0.91                  \\
UMAP n\_neighbors  & 30                   & 0.94                  \\
UMAP min\_dist     & 0.4                  & 0.94                 \\
\hline
\multicolumn{2}{l}{\textbf{Baseline}}              & 0.94                  \\
\hline
\end{tabular}
\caption{Robustness test results for the embedding model and UMAP parameters, with a concept size of $n=100$.}
\label{table:robustness-test}
\end{table*}

We conduct a robustness test to evaluate the impact of key parameters, including the embedding model and important UMAP hyperparameters. Details can be found in~\autoref{table:robustness-test}. We select \textit{stella\_en\_400M\_v5} as it is the current best-performing model (under 1B parameters) on MTEB Clustering.

For UMAP dimensionality reduction, we found that setting \textit{n\_components} beyond 3 was infeasible due to excessive memory requirements (exceeding 800GB for \textit{n\_components} = 4). Meanwhile, variations in \textit{n\_neighbors} and \textit{min\_dist} had minimal impact on performance.

\subsection{Pseudo Code}
\renewcommand{\algorithmicrequire}{\textbf{Input:}}
\renewcommand{\algorithmicensure}{\textbf{Output:}}
\begin{algorithm}
\caption{Pseudo Code for Blob Linking}
\label{code-blob-linking}
\begin{algorithmic}[1]
\REQUIRE List of blobs $B_{w}$ for each subsequent period following the given reference period
\ENSURE List of graphs $G_{b_{end}}$ of linked blobs for the given reference period, with $b_{end}$ being the ending blob of the graph
\FOR{each blob $b_{1,i} \in B_{1}$}
    \STATE Initialize $G_{b_{1,i}}$
\ENDFOR
\FOR{$w$ from $2$ to $W$}
    \FOR{each blob $b_{w,i} \in B_{w}$}
        \STATE Find blobs $b_{w-1,j} \in B_{w-1}$ with $dist(b_{w-1,j},b_{w,i}) < d$
        \FOR{each blob $b_{w-1,j}$ found}
            \STATE Add edge $(b_{w,i}, b_{w-1,j})$ to $G_{b_{w-1,j}}$
            \STATE Update the end of $G_{b_{w-1,j}}$ to $b_{p,i}$, thus $G_{b_{w,i}}$
        \ENDFOR
    \ENDFOR
\ENDFOR
\end{algorithmic}
\end{algorithm}

\subsection{Historical Event Selection}
\label{sec:event-selection}

Our initial event list was derived from relevant Wikipedia entries (\url{https://en.wikipedia.org/wiki/Outline_of_the_history_of_the_United_States}). We refined this list by excluding entries that were too broad (e.g., \textit{Patriotism}), primarily biographical (e.g., lists of U.S. presidents), or date-specific (e.g., war start/end dates).

The final selection includes:
World War I, Germania, Roaring Twenties, Great Depression, The New Deal, World War II, Cold War, Korean War, Assassination of President McKinley, Suez Crisis, Cuban Revolution, Civil Rights Movement, Brown v. Board of Education and "massive resistance," Vietnam War, Watergate, 1973 Oil Crisis, Reaganomics, and the Moon Landing.

\subsection{LLM Prompts for Sentence Generation and Summarization}
\label{section:appendix-prompt}
The prompt we use to generate sentences containing keywords for the synthetic dataset:

\textit{``You are an assistant who helps generate sentences containing a specific keyword. I will give you a keyword, and you will generate a list of sentences, each of which should contain the word \{KEYWORD\}, used according to its specified meaning: \{KEYWORD's NEW MEANING\}. The sentences should be written in the style of sentences in Corpus of Historical American English, such as \{5 COHA SENTENCES\}''}

The prompt we use to summarize the topics: 

\textit{``You are an assistant who is good at summarizing a list of texts. Read the list of texts and summarize them in at most 2 sentences, try to be as specific and detailed as possible. Remember all the texts in the list have to be closely related to your summarization. For example, if a list of texts are about "equal pay for women", your summarization needs to clearly mention that it is about equal pay for women, not just equal pay''}

\subsection{Concepts that were overrepresented by women senators}
\label{section:appendix-women}

\begin{itemize}
    \item \textbf{Electricity Market Manipulation and Regulatory Reform}: Broad discussions on preventing market manipulation in energy sectors and the need for stronger regulatory frameworks to protect consumers and ensure fair competition.
    \item \textbf{Sexual Assault in the Military}: Addressing systemic cultural issues and structural changes needed to improve how the military handles sexual assault cases and victim support.
    \item \textbf{Energy Deregulation and Consumer Impact}: Examination of the broader effects of energy market deregulation, with a focus on price stability, consumer protections, and the consequences of reduced oversight.
    \item \textbf{Firefighter Funding and Safety Standards}: General advocacy for increased support and funding for fire departments across the U.S., focusing on preparedness, training, and community safety.
    \item \textbf{Climate Change and Environmental Policy}: Legislative efforts to address climate change, focusing on balancing economic growth with environmental sustainability and national security.
    \item \textbf{Student Loan Debt and Higher Education Accessibility}: The growing crisis of student loan debt in the U.S., its economic impact, and strategies to make higher education more affordable and accessible.
    \item \textbf{Prescription Drug Costs and Healthcare Accessibility}: Broader issues around prescription drug pricing, the economic burden on consumers, and the need for improved affordability and transparency in healthcare.
    \item \textbf{Homeland Security Funding Allocation}: Broader discussions on effective homeland security funding, emphasizing the need for risk-based distribution and the protection of critical infrastructure.
    \item \textbf{Forest Management and Wildfire Prevention}: Legislative focus on sustainable forest management practices to reduce the frequency and severity of wildfires, protect communities, and maintain healthy ecosystems.
    \item \textbf{Post-Disaster Recovery and Federal Response}: Evaluations of federal disaster response strategies, accountability in recovery efforts, and long-term support for rebuilding communities affected by natural disasters.
    \item \textbf{Port Security and National Safety Concerns}: Ensuring comprehensive port security measures in response to vulnerabilities in maritime transportation, focusing on inspection protocols and technology improvements.
    \item \textbf{Judicial Diversity and Federal Court Effectiveness}: The importance of maintaining a diverse judiciary and filling court vacancies to ensure effective and timely judicial processes.
    \item \textbf{Healthcare Access and Patient Rights}: General discussions around ensuring that healthcare decisions prioritize patient needs over profit, with emphasis on patient protections and healthcare equity.
    \item \textbf{Workforce Development and Economic Mobility}: Legislative focus on workforce training, education, and support for economically disadvantaged communities to enhance social mobility and reduce inequality.
    \item \textbf{Affordable Housing and Urban Development}: Broader topics around housing affordability, urban planning, and support for vulnerable populations in maintaining stable housing.
    \item \textbf{Consumer Protection Against Deceptive Practices}: Efforts to combat deceptive marketing and protect consumers from fraudulent schemes, focusing on transparency and accountability.
    \item \textbf{Federal Budget Priorities and Economic Stability}: Ongoing debates around sustainable federal budgeting, spending priorities, and the long-term economic impact of budgetary decisions.
    \item \textbf{Energy Security and Resource Independence}: Discussions around ensuring energy security, reducing reliance on foreign oil, and investing in renewable resources to build a resilient energy infrastructure.
    \item \textbf{Economic Support for Low-Income Communities}: Strategies to address poverty and economic inequality through targeted social programs, job creation, and educational opportunities.
    \item \textbf{National Security and Intelligence Oversight}: Broader discussions on improving intelligence gathering, interagency cooperation, and maintaining civil liberties while ensuring national security.
\end{itemize}

\subsection{Concepts that were overrepresented by minority senators}

\label{section:appendix-minority}

\begin{itemize}
    \item \textbf{Native American Sovereignty and Self-Governance}: Ongoing legislative discussions and policy proposals aimed at increasing tribal autonomy, particularly in criminal justice and healthcare, while reducing federal oversight to promote self-determination and cultural preservation.
    
    \item \textbf{Environmental Justice and Water Rights for Indigenous Communities}: Focus on resolving water rights disputes and ensuring environmental protections for Native American lands, highlighting the intersection of environmental conservation and tribal rights.

    \item \textbf{Recognition of Veterans' Contributions and Welfare}: Broad discussions around the improvement of veterans' healthcare and support systems, reflecting a national effort to recognize veterans' sacrifices and provide equitable services for all who served.

    \item \textbf{Federal Oversight and Reform in Native American Affairs}: Debates about restructuring the Bureau of Indian Affairs, emphasizing the need to shift control from federal agencies to tribes, promoting autonomy and local governance.
    
    \item \textbf{Civil Rights and Criminal Justice for Minority Communities}: Examination of the legal system's fairness, particularly in relation to the federal death penalty's application in minority and Native American communities, focusing on equal justice and civil rights.

    \item \textbf{Economic Development and Political Status of Puerto Rico}: A nuanced exploration of Puerto Rico’s political autonomy, economic initiatives, and U.S. influence, reflecting on broader themes of decolonization and self-governance.

    \item \textbf{Healthcare Equity for Marginalized Groups}: Comprehensive policy discussions aimed at addressing disparities in healthcare access and outcomes for Native Americans and other minority communities, advocating for culturally competent care.

    \item \textbf{Comprehensive Immigration Reform}: Efforts to balance security, economic needs, and the humane treatment of undocumented immigrants, emphasizing the challenges of creating a fair immigration system without exacerbating labor exploitation.

    \item \textbf{Advocacy for Educational Opportunities in Minority Communities}: Legislation and policy debates focused on improving educational resources, preserving cultural identity, and supporting minority students, including efforts to empower local control.

    \item \textbf{Environmental and Energy Policy Leadership}: Minority senators' involvement in crafting and promoting sustainable environmental policies, such as balancing resource management with economic development and community health.

    \item \textbf{National Infrastructure and Equitable Resource Distribution}: Broad discussions on the need for a fair and effective allocation of federal funds for infrastructure, with a focus on supporting both urban and rural development equitably.

    \item \textbf{Marine Resource Management and Oceanography}: Promotion of oceanographic research and resource management, emphasizing the strategic and economic importance of U.S. leadership in marine science and environmental stewardship.

    \item \textbf{Combating Hate Crimes and Promoting Community Safety}: Legislative actions aimed at addressing and preventing hate-motivated violence, emphasizing the need for robust data collection and community-based interventions.

    \item \textbf{Flood Control and Sustainable Water Management}: Proposals for long-term flood control strategies and comprehensive water management plans to prevent natural disasters and support sustainable development across affected regions.

    \item \textbf{Healthcare Access and Reproductive Rights}: Broader debates on reproductive health policies, focusing on the rights of low-income women and the implications of federal healthcare funding decisions on marginalized groups.

    \item \textbf{Tourism as an Economic Driver}: Recognition of tourism's role in economic development, particularly in states with high reliance on tourism, and efforts to promote the U.S. as a global leader in travel and hospitality.

    \item \textbf{Empowering Small Business Development}: Initiatives focused on reducing barriers and promoting economic opportunities for small businesses, particularly in underserved and minority communities, without targeting specific legislation.

    \item \textbf{Military and Defense Readjustments in Local Economies}: Discussions around the economic and social impact of military base closures on communities, advocating for policies to support local economies during defense downsizing.

    \item \textbf{Public Health Preparedness and National Security}: Thematic focus on enhancing public health infrastructure and preparedness to address bioterrorism and pandemics, emphasizing coordinated national strategies and inter-agency collaboration.

    \item \textbf{Advocacy for Comprehensive Civil Rights Protections}: Broader legislative themes centered on expanding civil rights protections, addressing discrimination in multiple areas such as employment, housing, and healthcare for underrepresented groups.
\end{itemize}

\subsection{Concepts that were overrepresented by Republican senators}
\label{section:appendix-republican}

\begin{itemize}
    \item \textbf{Intelligence and Military Relations}: Discussions focusing on U.S. foreign policy and defense strategies in regions such as the Middle East, Central America, and Taiwan, particularly concerning arms control and military alliances.
    
    \item \textbf{U.S.-Soviet Relations During the Cold War}: Strategic assessments and briefings on Soviet military capabilities and U.S. efforts to counteract Soviet influence globally.
    
    \item \textbf{U.S. Relations with China and Taiwan}: Congressional debates on U.S. foreign policy towards China and Taiwan, focusing on diplomatic recognition and military support.
    
    \item \textbf{Military Retirement and Procurement}: Legislative discussions on military benefits, including retirement policies and the procurement process for defense equipment.
    
    \item \textbf{Environmental Regulations and Resource Management}: Policy debates on the Clean Water Act, environmental conservation, and management of natural resources, including energy innovations.
    
    \item \textbf{Energy Policy and Radioactive Waste Management}: Hearings focused on energy supply, innovation, and managing radioactive waste, with emphasis on nuclear energy safety.
    
    \item \textbf{Native American Sovereignty and Self-Governance}: Ongoing legislative discussions aimed at increasing tribal autonomy, particularly in criminal justice and healthcare, promoting self-determination.
    
    \item \textbf{Economic Impacts of Rising Federal Debt}: Review of the growing U.S. federal debt and its long-term economic consequences, particularly after the debt exceeded \$5 trillion by 2000.
    
    \item \textbf{Hate Crimes and Violence Against Marginalized Groups}: Reports of rising hate crimes based on race, sexual orientation, and gender identity, sparking debates on the need for stronger hate crime legislation.
    
    \item \textbf{Oversight of National Security Policies}: Evaluations of U.S. national security and defense policies, with particular focus on international conflicts such as the invasion of Grenada.
    
    \item \textbf{Military Appointments and Honors}: Discussions and acknowledgments of military appointments, promotions, and the recognition of veterans' contributions across multiple branches of the U.S. military.
    
    \item \textbf{Trade Relations with the European Community}: Congressional hearings on U.S. trade policies with European nations, focusing on economic competition, tariffs, and diplomatic relations.
    
    \item \textbf{Marshall Plan and Post-War Trade Policies}: Debates on the impact of the Marshall Plan and post-WWII U.S. foreign policy, particularly concerning trade relations with Eastern Europe.
    
    \item \textbf{Judicial Activism and Marriage Laws}: Legislative responses to judicial rulings on marriage, particularly surrounding the definition of marriage and the role of federal versus state authority.
    
    \item \textbf{U.S. Involvement in the Korean War}: Debates on U.S. military intervention in Korea, focusing on constitutional authority, military strategy, and the broader implications for U.S. foreign policy.
    
    \item \textbf{Foot-and-Mouth Disease (FMD) in Livestock}: Concerns over the threat of FMD outbreaks in neighboring countries, leading to discussions on U.S. agricultural biosecurity and disease prevention measures.
    
    \item \textbf{NATO and U.S. Military Commitments}: Analysis of U.S. military obligations under NATO, with debates on the potential risks of entanglement in European conflicts during the Cold War.
    
    \item \textbf{Tribal Sovereignty and Federal Law}: Legislative debates on tribal sovereignty, focusing on the application of federal death penalty laws on Native American reservations.
    
    \item \textbf{U.S.-Spain Relations During the Cold War}: Congressional discussions on U.S. military alliances and the strategic importance of Spain in the broader NATO defense framework.
    
    \item \textbf{Debt Ceiling and Fiscal Responsibility}: Ongoing debates over raising the U.S. debt ceiling, with emphasis on fiscal responsibility, government spending, and the risk of financial crises.
\end{itemize}

\subsection{Concepts that were overrepresented by Democratic senators}
\label{section:appendix-democrat}

\begin{itemize}
    \item \textbf{Immigration and Naturalization Laws}: Legislative efforts to provide exemptions, waivers, and status adjustments, particularly for family members of U.S. citizens, reflecting a trend towards facilitating family reunification and humanitarian considerations.
    
    \item \textbf{Indian Affairs and Tribal Legislation}: Public hearings and legislative meetings focused on Indian affairs, including land claims, healthcare, housing, and tribal recognition, indicating ongoing efforts to address Native American concerns.
    
    \item \textbf{Transportation and Science Oversight}: Senate Committee hearings on topics like transportation safety, telecommunications, and environmental impacts, with a focus on federal regulation of transportation industries during the late 1970s and 1980s.
    
    \item \textbf{Environmental and Energy Policy Hearings}: Discussions on environmental legislation, such as the Clean Air Act, nuclear waste management, and global climate change, reflecting legislative efforts to address pressing environmental challenges.
    
    \item \textbf{Agricultural Policy and Food Security}: Hearings focused on agricultural policy, including preparations for farm bills, water quality, and global warming's impact on agriculture, emphasizing the Senate's role in shaping food security and agricultural sustainability.
    
    \item \textbf{Women's Issues and Economic Policy}: Senate hearings on a wide array of topics, including workplace discrimination, mortgage lending, healthcare, and education, reflecting legislative efforts to address social and economic challenges affecting women.
    
    \item \textbf{Native American Health and Environmental Legislation}: Hearings on Native American healthcare and environmental policies, including the reauthorization of the Indian Health Care Improvement Act, highlighting federal responsibilities toward Native communities.
    
    \item \textbf{Small Business Protections and Regulatory Challenges}: Senate Select Committee hearings on issues affecting small businesses, such as regulatory barriers, financial assistance programs, and the economic impact of federal policies.
    
    \item \textbf{Civil Rights and Voting Rights Legislation}: Legislative debates on civil rights and voting laws, addressing systemic disenfranchisement and discriminatory practices, with a focus on increasing protections for marginalized groups.
    
    \item \textbf{Genocide Convention and Human Rights Legislation}: Advocacy for the ratification of the United Nations Genocide Convention, emphasizing the U.S. commitment to human rights and moral leadership in preventing atrocities.
    
    \item \textbf{Drug Pricing and Pharmaceutical Regulations}: Legislative efforts to address drug pricing, focusing on the cost disparity between generic and branded drugs, and the push for increased transparency and competition in the pharmaceutical industry.
    
    \item \textbf{Fishing Industry Legislation and Foreign Competition}: Legislative discussions on protecting the U.S. fishing industry from foreign competition, including subsidies for American fishermen and conservation practices to sustain marine resources.
    
    \item \textbf{Environmental Conservation and National Parks Legislation}: Hearings on the establishment of national seashores and parks, such as the Oregon Dunes and Indiana Dunes, with a focus on environmental preservation and public access.
    
    \item \textbf{Healthcare for the Elderly}: Legislative efforts to provide better health insurance and financial assistance for elderly citizens, with proposals such as the Anderson amendment to integrate medical care into the social security system.
    
    \item \textbf{Foreign Aid and Developmental Assistance}: Debates on reforming U.S. foreign assistance programs, with emphasis on efficiency, accountability, and aligning aid with U.S. interests while promoting development in recipient countries.
    
    \item \textbf{Urban Development and Housing Policy}: Hearings on urban revitalization, housing finance reform, and addressing the impacts of financial crises on housing markets, with a focus on providing affordable housing solutions.
    
    \item \textbf{Nuclear Weapons and Arms Control}: Discussions on U.S. nuclear policies, arms control agreements, and efforts to suspend nuclear weapons testing, emphasizing the need for international cooperation and inspection systems to ensure global security.
    
    \item \textbf{Forest Management and Conservation}: Debates on forest conservation policies, timber management, and the need for sustainable forestry practices to protect national forests and promote economic growth in forest-dependent communities.
    
    \item \textbf{Aircraft Noise Pollution and Environmental Impact}: Legislative proposals to address the negative impact of aircraft noise on communities, advocating for noise control measures, quieter engine technology, and public health protections.
    
    \item \textbf{School Lunch Programs and Child Nutrition}: Legislative efforts to address child hunger through the National School Lunch Program, including extending food assistance during summer months and maintaining food security for low-income children.
\end{itemize}

\end{document}